\documentclass{article}

\usepackage[final]{neurips_2019}

\usepackage[utf8]{inputenc} 
\usepackage[T1]{fontenc}    
\usepackage{hyperref}       
\usepackage{url}            
\usepackage{booktabs}       
\usepackage{amsfonts}       
\usepackage{nicefrac}       
\usepackage{microtype}      

\usepackage{graphicx} 
\usepackage{float} 
\usepackage{subfigure} 
\usepackage{amsmath}
\usepackage{color}
\usepackage{array}
\usepackage{xspace}

\usepackage{amsmath,amsfonts,amssymb,bm}
\usepackage{pifont} 
\usepackage{graphicx,subfigure,epsfig,fancybox} 
\usepackage{float}
\usepackage{color} 
\usepackage{multirow}
\usepackage{natbib}

\newcommand{\revised}[1]{\textcolor{blue}{}}
\renewcommand{\vec}[1]{\boldsymbol{#1}}

\newcommand{\mat}[1]{\mathbf{#1}}
\newcommand{\trans}[1]{#1^{\textsf{T}}}

\DeclareMathOperator*{\argmin}{argmin}
\DeclareMathOperator*{\argmax}{argmax}

\usepackage{adjustbox}
\usepackage{array}
\usepackage{booktabs}

\newcolumntype{R}[2]{%
    >{\adjustbox{angle=#1,lap=\width-(#2)}\bgroup}%
    l%
    <{\egroup}%
}

\renewcommand{\trans}[1]{\ensuremath{#1^{\top}}}

\usepackage{tikz}
\usetikzlibrary{shapes.geometric, positioning, calc}
\usetikzlibrary{arrows,shapes,calc}
\usetikzlibrary{trees,positioning,mindmap,shadows,fit}
\usetikzlibrary{decorations.pathreplacing}
\usetikzlibrary{tikzmark} 
\usetikzlibrary{intersections} 

\newcommand{\newmodel}{\textsc{Cnn-Ek}\xspace}
\newcommand{\newmodeladv}{\textsc{Cnn-Adv}\xspace}
\newcommand{\oldmodel}{\textsc{Cnn}\xspace}
\newcommand{\newrnn}{\textsc{Rnn-Ek}\xspace}
\newcommand{\oldrnn}{\textsc{Rnn}\xspace}
\newcommand{\daek}{\textsc{Da-Ek}\xspace}
\newcommand{\daadv}{\textsc{DA-Adv}\xspace}

\title{Improving the Explainability of Neural Sentiment Classifiers via Data Augmentation}

\author{Hanjie Chen \\
	Department of Computer Science \\
	University of Virginia \\
	Charlottesville, VA 22904 \\
	\texttt{hc9mx@virginia.edu} \\\And
	Yangfeng Ji \\
	Department of Computer Science \\
	University of Virginia \\
	Charlottesville, VA 22904 \\
	\texttt{yangfeng@virginia.edu} \\}

\begin{document}
\maketitle
\begin{abstract}
  Sentiment analysis has been widely used by businesses for social media opinion mining, especially in the financial services industry, where customers' feedbacks are critical for companies. Recent progress of neural network models has achieved remarkable performance on sentiment classification, while the lack of classification interpretation may raise the trustworthy and many other issues in practice.
In this work, we study the problem of improving the explainability of \emph{existing} sentiment classifiers.
We propose two data augmentation methods that create additional training examples to help improve model explainability: one method with a predefined sentiment word list as external knowledge and the other with adversarial examples.
We test the proposed methods on both CNN and RNN classifiers with three benchmark sentiment datasets.
The model explainability is assessed by both human evaluators and a simple automatic evaluation measurement.
Experiments show the proposed data augmentation methods significantly improve the explainability of both neural classifiers.
\end{abstract}

\section{Introduction}
\label{sec:intro}
Sentiment analysis is one of the most widely-used applications of natural language processing (NLP) in the financial services industry~\citep{sohangir2018big, daniel2017company, li2018transformation, chen2016neural}, where neural sentiment classifiers help enterprises gauge public opinion, conduct market research, monitor brand and product reputation, and understand customer experiences~\citep{feldman2013techniques,pang2002thumbs,liu2012sentiment}. 
The recent development of neural network modeling has largely boosted the prediction performance (e.g., accuracy) on sentiment classification~\citep{yang2016hierarchical, johnson2017deep, johnson2016supervised, zhou2016text}, while the nonlinearity of neural network models hinders the understanding on predictions.
A fair question with no easy answer for neural sentiment classifiers is \emph{why the prediction on this text is positive (or negative)?}
Moreover, the lack of explainability on model prediction will raise the issue of trustworthy and fairness of sentiment classifiers in practice~\citep{gilpin2018explaining,yu2019three}.

\begin{table}
  \centering
  \begin{tabular}{rp{0.35\textwidth}}
    \toprule
    Review & a truly moving experience, and a perfect example of how art when done right can help heal, clarify, and comfort.\\
    \midrule
    Pred. A & Positive \\
    Exp. A & a, moving, can\\
    \midrule
    Pred. B & Positive\\
    Exp. B & perfect, comfort, truly\\
    \bottomrule
  \end{tabular}
  \caption{Explanations generated by the local explanation method \textsc{LIME} from two neural sentiment classifiers. The ground-truth sentiment polarity of the text is positive and \emph{both} models give the right prediction.}
  \label{tab:intro-example}
\end{table}

To address the explainability issue of neural classifiers, various approaches have been developed recently to provide model-agnostic explanations on predictions~\citep{ribeiro2016should, ribeiro2018anchors, lundberg2017unified}.
Particularly, this work focuses on local explanations, which aims to explain predictions for individual data.
The most common way of generating local explanations in sentiment classification is identifying the important part of a text associated with predicted sentiment polarity~\citep{lei2016rationalizing,chen2018learning,nguyen2018comparing}.
For example, a widely-adopted local explanation method \textsc{LIME}~\citep{ribeiro2016should} can identify a set of keywords as an explanation.\footnote{A brief description of \textsc{LIME} is provided in \autoref{sec:setup}.} 

\autoref{tab:intro-example} presents an example of movie reviews and the explanations based on two neural sentiment classifiers.
Although both classifiers give the right prediction on this example, the explanation A is harder to be \emph{interpreted} than the explanation B, in terms of why the prediction is positive.
This difference on the explainability of explanations leads us to trust more on prediction B than A, which will eventually discriminate the practical values of these two sentiment classifiers. 

In general, a prediction explanation can be generated by using any local explanation method \citep{ribeiro2016should, kindermans2017learning, fong2017interpretable, dabkowski2017real}.
However, the real challenge in practice is that whether an explanation is easy to be interpreted, as demonstrated in \autoref{tab:intro-example}.
By noticing the connection and difference between explanations and their explainability, we would like to study the problem on {\bf improving the explainability of neural sentiment classifiers}.
We consider this as a learning problem and propose to resolve it with some data augmentation methods.
The goal is to increase the explainability of existing neural sentiment classifiers while maintaining similar prediction performance. 

In this work, we explore the strategy of using data augmentation to improve the explainability of neural sentiment classifiers. 
We propose two data augmentation methods: one with a predefined sentiment word list as external knowledge and the other with adversarial examples.
Experiments on the two base models and three benchmark datasets show that the proposed methods improve the model explainability with respect to both automatic and human evaluation.

\section{Data Augmentation Methods}
\label{sec:methods}

The basic idea is to teach the models to make predictions based on critical information.
In the scenario of sentiment classification, the task is to teach model to make predictions by grasping sentiment words.
This section presents two proposed methods for data augmentation and a unified method of using augmented data for training. 

\subsection{Augmenting via External Knowledge}
\label{subsec:methods-external}

The first method is called data augmentation with external knowledge (\daek).
We propose a simple method to create some examples that are similar to training examples with respect to their surface forms, but those examples do not belong to any of the predefined classes $\mathcal{Y}$.
To be specific, the augmented examples for sentiment analysis are the examples that are similar to original training examples but have no sentiment polarity, as illustrated in \autoref{tab:da-examples}.

A simple way to create an augmented example $\tilde{\vec{x}}$ is that, for a given sentence $\vec{x}$, removing words $\{\vec{x}_i\}$ from $\vec{x}$ if $\vec{x}_i$ belongs to a predefined sentiment word list.
In this work, we use the words listed in the SentiWordNet corpus \citep{baccianella2010sentiwordnet} and their sentiment polarity scores.
For a given sentence $\vec{x}$, removing $\vec{x}_i$ from $\vec{x}$ if $\vec{x}_i$ is in the word list will create an augmented example $\tilde{\vec{x}}$.
\autoref{tab:da-examples} presents two examples of the original text and its augmented counterpart after removing words with clear sentiment polarity.
For some simple texts, removing sentiment words will cause their augmented counterparts to be incomplete sentences, which can still be used as augmented data points.
For example, if we remove the sentiment word in text \texttt{I like this movie}, then the augmented training example is \texttt{I this movie}.
However, with the training framework proposed in \autoref{subsec:methodslearning}, this augmented example will help the model to emphasize the sentiment prediction on the original sentence.

There is a critical distinction between the augmented examples created by \daek and the example from the \emph{neutral} class in sentiment classification.
In multi-class sentiment classification, there is often a class with an average sentiment score called the neutral class.
The major difference is that texts from a neutral class still have sentiment, or at least contains some sentiment words. 
For example, the movie review \texttt{The Cockettes provides a window into a subculture hell-bent on expressing itself in every way imaginable.} is a neutral but not augmented example.
With words like \texttt{hell-bent} and \texttt{imaginable}, it shows sentiment inclination of this movie review even though it is not strong.
To construct an augmented example from this text, the proposed method still needs to remove the sentiment words.
Empirical results show that adding neutral examples can only lead to a minor improvement on explainability.

\subsection{Augmenting with Adversarial Examples}
\label{subsec:methods-adversarial}

This method is called data augmentation with adversarial examples or \daadv.
We adopt the method proposed by \citet{alzantot2018generating} to generate adversarial examples, which may have similar surface forms and semantic meanings to training examples. 
To be specific, this method aims to minimize the number of modified words between the original and adversarial examples, and maintain semantic and syntactic similarity by substituting only a few synonyms. 
A well-known challenge on generating adversarial examples in text data is that texts are discrete, which causes the difficulty in generating adversarial examples by using the popular gradient-based methods \citep{goodfellow2014explaining, kurakin2016adversarial, madry2017towards}. 
\citet{alzantot2018generating} developed an attack algorithm via genetic algorithms. 
In each generation, a group of candidate examples are generated by substituting synonyms, and those most fit within the context surrounding are selected by the Google 1-billion words language model~\citep{chelba2013one}. 
The candidates that can successfully attack the model to flip prediction polarity are adversarial examples. 
Like many other adversarial attack methods, there is a budget about how many words can be replaced. 
Beyond that budget limit will cause a fail attack. 
In our case, it means not every text can get an adversarial example.

As adversarial examples can flip model predictions, the replaced words from original texts must be critical to sentiment prediction.
Similar to the previous data augmentation method, we can construct augmented examples by taking the replaced words as the sentiment words in \daek. 
 
\paragraph{Comparison.}
\autoref{tab:da-examples} presents some examples of two data augmentation methods. 
With \daek, we have a high-precision method for data augmentation.
If any word in a text matches one entry in the SentiWordNet, then it is very likely to be a sentiment word.
However, the word list in the SentiWordNet is predefined and definitely not comprehensive.
The missing sentiment words imply \daek could be a data augmentation method with low recall.
With \daadv, we have a low-precision method for data augmentation.
Words identified by adversarial attacks can be sentiment words or simply can be non-sentiment words that are sensitive to neural sentiment classifiers.
Besides, finding adversarial examples is very time consuming, as further explained in \autoref{sec:exp}.
But \daadv has the potential to extend this method to other text classification tasks, where we do not have a pre-defined word list.

\begin{table*}
  \centering
  {
    \begin{tabular}{rp{0.6\textwidth}}
      \toprule
      Original text & the only problem is that, by the end, no one in the audience or the film seems to really care \\
      \daek & the only that , by the end, one in the audience or the film seems to care \\[0.3em]
      Adversarial example & the only difficulty is that, by the end, no one in the audience or the movie seems to really caring \\
      \daadv & the only is that, by the end, no one in the audience or the seems to really  \\
      \midrule
      Original text & michel piccoli's moving performance is this films reason for being \\
      \daek & michel piccoli's this films reason for being \\[0.3em]
      Adversarial example & michel piccoli's moving play is this movie reason for being \\
      \daadv & michel piccoli's moving is this reason for being \\
      \bottomrule
    \end{tabular}
    \caption{\label{tab:da-examples}Some examples of the augmented data created by \daek and \daadv.}
  }
\end{table*}

\subsection{Learning with Augmented Examples}
\label{subsec:methodslearning}

We extend the training set $\mathcal{D} = \{(\vec{x}^{(k)},y^{(k)})\}$ as by adding the augmented examples $\{(\tilde{\vec{x}}^{(k')},\textsc{aug})\}$ generated by either \daek or \daadv and extend it as $\widetilde{\mathcal{D}} = \mathcal{D}\cup\{(\tilde{\vec{x}}^{(k')},\textsc{aug})\}$. 
Similarly, the label set $\mathcal{Y}$ is also extended as $\widetilde{\mathcal{Y}}=\mathcal{Y}\cup\{\textsc{aug}\}$.
Note that, the proposed methods only create augmented examples for the training set and development set.
No modification is on the test set.

Once we have the extended training and development set, learning a neural sentiment classifier with data augmentation is straightforward.
Specifically, we optimize the following loss function
\begin{equation}
  \label{eq:obj}
  \argmin_{\vec{\theta}}\sum_{\widetilde{\mathcal{D}}}\mathcal{L}(\hat{y}^{(k)},y^{(k)}),
\end{equation}
to achieve the best prediction accuracy on the \emph{augmented} development set,
where $\hat{y}^{(k)}$ is decoded from the decision function defined in \autoref{eq:cnn-decision} with extended label set $\widetilde{\mathcal{Y}}$, $\mathcal{L}(\cdot,\cdot)$ is the cross-entropy loss, and $\vec{\theta}$ is the collection of parameters, which is the same as the base model.
During test, with no augmented example, the trained neural classifier simply ignores any prediction on the label \textsc{aug} and picks the label from $\mathcal{Y}$ that maximizes the decision value in \autoref{eq:cnn-decision}. 
More information of the neural classifiers used in our experiments is provided in \autoref{sec:other}.

\section{Experimental Setup}
\label{sec:setup}

This section describes the experimental setup used in this work.
We test the proposed data augmentation methods with two neural sentiment classifiers, a convolutional neural network in \citep[\oldmodel]{kim2014convolutional} and a recurrent neural network with LSTM \citep[\oldrnn]{hochreiter1997long}, on three benchmark datasets, SST \citep{socher2013recursive}, MR \citep{pang2005seeing}, and IMDB \citep{maas2011learning}. 
Local explanations were generated from  LIME~\citep{ribeiro2016should} with model predictions and the cosine similarity method based on text representations. 
More implementation details are in Appendix~\ref{sec:other}. 

\subsection{Local Explanation Generation}
\label{subsec:explanation}

To generate local explanations, we adopt the LIME proposed by \citet{ribeiro2016should} to generate explanations on model predictions.
Besides, we also suggest another way of generating local explanations based on the cosine similarity between word representations and text representations.

\subsubsection{LIME with model predictions}

The basic idea of the LIME is that, for a given example $\vec{x}$, it finds an explanation based on a locally linear approximation $\vec{g}(\vec{z}^{(l)},y)$ of the decision function $\vec{h}(\vec{z},y)$, in which $\vec{z}$ is a perturbation of $\vec{x}$ obtained by subsampling the words from $\vec{x}$.
Given a set of subsamples $\{\vec{z}^{(l)}\}$ from $\vec{x}$, the loss function of the LIME is defined as 
\begin{equation}
  \label{eq:lime}
  L(\vec{h}, \vec{g})=\sum_{i=1}^{q} D_{\vec{x},\vec{z}^{(l)}}(\vec{h}(\vec{z}^{(l)},y)-\vec{g}(\vec{z}^{(l)},y))^2,
\end{equation}
where the linear approximation function $\vec{g}$ is usually defined as $\vec{g}(\vec{z},y) = \trans{\vec{w}_y}\vec{z}$.
$D_{\vec{x},\vec{z}^{(l)}}$ measures the similarity between $\vec{x}$ and $\vec{z}^{(l)}$,
\begin{equation}
  \label{eq:d-xz}
  D_{\vec{x},\vec{z}^{(l)}}=\exp\left(\frac{-d(\vec{f}(\vec{x}), \vec{f}(\vec{z}^{(l)}))^{2}}{\sigma^{2}}\right),
\end{equation}
with $d(\vec{f}(\vec{x}), \vec{f}(\vec{z}^{(l)})$ as the cosine distance between the latent representations of $\vec{x}$ and $\vec{z}^{(l)}$, as suggested in \citep{ribeiro2016should}.

Optimizing \autoref{eq:lime} will try to match the decision values from linear approximation $\vec{g}(\vec{z}^{(l)},y)$ with $\vec{h}(\vec{z}^{(l)},y)$ and also produce a set of linear weights $\{\vec{w}_y\}_{y\in\mathcal{Y}}$ associated with $\mathcal{Y}$.
The values of $\{\vec{w}_{y,i}\}$ indicate the importance of $\{\vec{x}_i\}$.
If the predicted label is $\hat{y}$, then top $t$ words according to $\{\vec{w}_{\hat{y},i}\}$ will be selected as an explanation of $\vec{x}$ on the corresponding prediction.

\subsubsection{Cosine similarity on text representations}
For a given text $\vec{x}$, the basic idea of using cosine similarity generating explanations is to measure the similarity between its text representation $\vec{f}(\vec{x})$ and word representations $\vec{f}(\vec{x}_i)$, where $\vec{x}_i$ is the embedding of the $i$-th word in text $\vec{x}$.
In this way, we can find the most similar words with respect to the text representation $\vec{f}(\vec{x})$, and choose the top $t$ words as an explanation.
The underlying assumption of this idea is that, if a text representation $\vec{f}(\vec{x})$ could facilitate sentiment prediction, the sentiment polarity indicated by the top $t$ similar words should be consistent with its overall sentiment polarity. 

To compute cosine similarity, we first need to map all the word embeddings $\{\vec{x}_1,\dots,\vec{x}_n\}$ into the text representation space using $\vec{f}(\cdot)$.
Then, the similarity between a text and the $i$-th word within this text is measured by the cosine value of these two vectors,
\begin{equation}
  \label{eq:cos-sim}
  \text{cos-sim}(\vec{f}(\vec{x}), \vec{f}(\vec{x}_i)) = \frac{\langle \vec{f}(\vec{x}),\vec{f}(\vec{x}_i)\rangle}{\|\vec{f}(\vec{x})\|_2\cdot\|\vec{f}(\vec{x}_i)\|_2}.
\end{equation}
After applying \autoref{eq:cos-sim} to every word in text $\vec{x}$, then we pick the top $t$ words with respect to their cosine similarities as an explanation of $\vec{f}(\vec{x})$.

\section{Experiments}
\label{sec:exp}

Although there are different ways to evaluate prediction explanations as suggested in prior work~\citep{ribeiro2016should,gilpin2018explaining}, the explainability of explanations should be the most of important criterion.
As argued by \citet{gilpin2018explaining}, a good explanation should be easily interpretable and ``\emph{simple enough for a person to understand using a vocabulary that is meaningful to the user}''.
For sentiment classification, as demonstrated in the running example (\autoref{tab:intro-example}), a good explanation on sentiment prediction should be easy enough for a human user to understand together with the prediction.
Following this intuition, we define the explainability measurement for both automatic evaluation and human evaluation.

\subsection{Automatic Evaluation}
\label{subsec:automatic}

Our automatic evaluation method measures the explainability of an local explanation (consisting of a set of keywords) by predicting its sentiment polarity and comparing with the model prediction.
Specifically, for each keyword in an explanation, we retrieve its sentiment scores from the SentiWordNet.
SentiWordNet offers three scores for a sentiment word: a positive score, a negative score, and a neutral score.
For the word \texttt{truly}, its positive score is 0.625, negative score is 0 and neutral score is 0.375, which indicates that it is a word with positive polarity.
On the other hand, the positive score of word \texttt{a} is 0 and its neutral score is 1. 
With the sentiment scores of these words, the overall scores of an explanation is the accumulation of the sentiment scores of its keywords.

Consider the local explanations in \autoref{tab:intro-example}, the sentiment scores of explanation A with respect to the positive sentiment polarity is 0 and the positive socre of explanation B is 0.625.
Under this simple automatic evaluation measurement, the explanation A is easier to be interpreted than explanation B, which is consistent with our expectation. 

To quantificationally evaluate prediction explanations, we propose the \emph{coherence score} defined as follow:
for a given test example, depending on the sentiment polarities predicted by the model, indicated by the explanation and the ground truth, it will be counted as a coherent case, if it satisfies one of the two conditions:
\begin{itemize}
\item Condition 1: if the sentiment polarity indicated by the explanation is not \textsc{none} and is consistent with the model prediction; or
\item Condition 2: if the sentiment polarity indicated by the explanation is \textsc{none} and the model prediction is not the same as the ground truth. 
\end{itemize}
The coherence between the model prediction and its prediction in condition 1 is obvious.
About condition 2, we consider that an explanation with no sentiment polarity is also coherent with a wrong prediction.
Intuitively, an explanation with no sentiment polarity explains why the prediction is wrong. 
For a collection of explanations, the coherence score is the ratio of the number of coherent cases to the total number of instances. 

\subsubsection{Results}


\begin{table*}
  \centering
  \begin{tabular}{p{0.15\textwidth}p{0.2\textwidth}p{0.1\textwidth}cc}
    \toprule
    & & & \multicolumn{2}{c}{Coherence}\\
    \cmidrule(lr){4-5}
    Dataset & Model & Accuracy & ~~~LIME~~~ & ~Cos-Sim~ \\
    \midrule
    SST & \oldmodel & 0.85  & 0.65 & 0.58 \\
            & \newmodel & 0.85  & 0.70 & 0.60 \\
            & \newmodeladv & 0.85  & 0.68 & 0.59 \\
    \cmidrule(lr){2-5}
            & \oldrnn & 0.84 &  0.64 & 0.61 \\
            & \newrnn & 0.84  & 0.66 & 0.62 \\
    \midrule
    MR & \oldmodel & 0.81 & 0.63 & 0.50  \\
    & \newmodel & 0.80 & 0.66 & 0.55  \\
    & \newmodeladv & 0.80  & 0.65 & 0.55 \\
    \cmidrule(lr){2-5}
    & \oldrnn & 0.80 & 0.64 & 0.59  \\
    & \newrnn & 0.80 & 0.65 & 0.60  \\
    \midrule
    IMDB & \oldmodel & 0.90 & 0.76 & 0.23  \\
    & \newmodel & 0.90 & 0.80 & 0.53 \\
    & \newmodeladv & 0.90  & 0.78 & 0.47 \\
    \cmidrule(lr){2-5}
    & \oldrnn & 0.87 & 0.78 & 0.74 \\
    & \newrnn & 0.87  & 0.81 & 0.78 \\
    \bottomrule
  \end{tabular}
  \caption{The classification and explainability evaluation results of different models on SST, MR and IMDB. The models trained with augmented data from \daek are named with \textsc{-Ek}. The model trained with the augmented data from \daadv is named with \textsc{-Adv}.}
  \label{tab:results}
\end{table*}

\autoref{tab:results} shows the prediction accuracies and coherence scores of different models on the all three datasets.
As indicated in the third column, the models trained with augmented data, including both \daek and \daadv, maintain the prediction accuracies comparing to their counterparts.
This observation matches our expectation that data augmentation for improving explainability should not hurt prediction accuracy.

More important, all models trained with augmented data outperform the base models with respect to the coherence score (column 4 and 5).
Comparing the coherence scores with the same base model and the same dataset, we found that, in most of the cases, both data augmentation methods help improve the coherence score, regardless which explanation generation we use.
Comparing the scores across multiple datasets and models, we also notice that the improvement on LIME-based explanations has a smaller variance than the explanations generated by the cosine similarity method.
We suspect that this is because local explanations are always tied with model predictions, while the cosine similarity method only use text representations to generate explanations. 

As shown in the experiments with the \newmodeladv, data augmentation with adversarial examples does provide some benefit to improve the coherence scores of the \oldmodel model on all of the three datasets. The state-of-the-art method \citep{alzantot2018generating} generating adversarial examples can be extended to other neural calssifiers (e.g. \oldrnn) and text classification tasks in the future work.

\subsection{Human Evaluation}

In \autoref{subsec:automatic}, we propose the coherence score and use it to automatically evaluate the local explanations.
Even though it is easy to compute, the major limitation is from the pre-defined list of sentiment words.
For a specific test example, this evaluation method will fail if the sentiment words in the generated explanation are not the SentiWordNet word list. 
Furthermore, as discussed in the beginning of this section, explainability is about whether an explanation is understandable to human users.
Human evaluation is necessary if we would like to measure the explainability improvement. 
Besides the human evaluation results can provide further justification of the coherence scores from automatic evaluation.

\subsubsection{Evaluation setup}

To conduct a human evaluation task, we random pick 100 test examples from the SST and MR datasets.
Explanations of these examples are generated by LIME based on the \oldmodel and \newmodel models. 
We have 7 graduate students with proficient English skills as volunteers to evaluate the quality of these explanations.

With a given test example with an explanation pair generated from the \oldmodel and \newmodel models respectively, a human evaluator needs to a two-step evaluation.
First, for each explanation, the human evaluator needs to analyze whether it can interpret the corresponding prediction, and mark with a  score ("1" for coherent, "0" for incoherent) according to the two conditions in \autoref{subsec:automatic}, only with the evaluator himself to give the sentiment polarity of the explanation.
Then, for the explanation pair, the evaluator will be asked to pick which one better explains the corresponding model prediction. 
Note that, two explanations within each pair are presented to our human evaluators randomly to eliminate any possible bias.

Finally, the human evaluation score is calculated as the ratio of the sum of the scores to the number of examples. 
We also calculate the coherence scores on the 100 test examples and compare them with the human evaluation scores.

\subsubsection{Results}


\begin{table*}
  \centering
  \begin{tabular}{llcc}
    \toprule
    Dataset & Model & Human Evaluation & Automatic Evaluation\\
    \midrule
    SST & \oldmodel & 0.85 & 0.56\\
            & \newmodel & 0.92 & 0.63\\
    \midrule
    MR & \oldmodel & 0.84 & 0.55\\
            & \newmodel & 0.90 & 0.60\\
    \bottomrule
  \end{tabular}
  \caption{Human and automatic evaluation results on the sets of the SST and MR datasets. The augmented examples are from \daek and the explanations are generated by LIME.}
  \label{tab:human-eval}
\end{table*}

\autoref{tab:human-eval} presents both the human evaluation scores and also the coherence scores.
On both datasets, human evaluation scores indicate that data augmentation with additional examples improves the explainability of \oldmodel.
As shown in \autoref{tab:examples}, the explanations from \newmodel are more interpretable and the sentiment polarity indicated by these two explanations are clear. 
In addition, the comparison between the human evaluation and the automatic evaluation also shows the coherence scores are positively correlated with the human evaluation scores, which provides a justification for our automatic evaluation measurement.

We also notice that the coherence scores are constantly lower than the human evaluation scores, even though the computations of these two scores are similar.
One possible reason is that the pre-defined sentiment word list from the SentiWordNet is not comprehensive enough, while human evaluators can always tell which explanation is better than the other. 

\begin{table*}
  \renewcommand\arraystretch{2}
  \small
  \centering
  \begin{tabular}{p{0.3\textwidth}cccc}
    \toprule
    Input text & Models & Prediction & Keywords & Indication \\
    \midrule
    \multirow{2}{*}{\parbox{0.3\textwidth}{\textcolor{red}{[Pos]} at about 95 minutes, treasure planet maintains a brisk pace as it races through the familiar story}}
               & \oldmodel & Positive & treasure, pace, a & None \\
    \cmidrule{2-5} 
               & \newmodel & Positive & brisk, treasure, familiar & Positive \\
    \midrule
    \multirow{2}{*}{\parbox{0.3\textwidth}{\textcolor{blue}{[Neg]}
    unfortunately, they're sandwiched in between the most impossibly dry account of kahlo's life imaginable}}
               & \oldmodel & Negative & dry, 're, sandwiched & None \\
    \cmidrule{2-5} 
               & \newmodel & Negative & unfortunately, dry, account & Negative \\
    \bottomrule
  \end{tabular}
  \caption{Examples of the explanations generated by LIME from the \oldmodel and \newmodel models, where the ground truth of each input text is marked in front of it as "Pos" or "Neg".}
  \label{tab:examples}
\end{table*}

\section{Related Work}
\label{sec:related}

Deep learning has demonstrated success on text classification, including sentiment analysis.
Along with the increase of neural network models, there is a growing demanding on building explainable models~\citep{gilpin2018explaining,guidotti2018survey,murdoch2019interpretable}.
Particularly in sentiment classification, prior work on recommendation systems with sentiment analysis shows that recommendations with explanations have more influential impact on users behavior~\citep{zhang2014explicit,zhang2015incorporating}.
Similar work has also been shown in other application domain of sentiment analysis.
For example, \citet{luo2018beyond} demonstrate an explainable neural network model can uncover informative clues related users preference in financial sentiment analysis.

This work focus on local explanations, which are generated based on individual model predictions.
Among the local explanation methods, LIME \citep{ribeiro2016should} is probably the most popularly used method, due to its model-agnostic feature.
Some other model agnostic explanation methods include SHAP~\citep{lundberg2017unified} and MAPLE~\citep{lundberg2017unified}, which we will leave for future work.
Besides, there are model-based local explanation methods, for example, using saliency maps from gradient information~\citep{li2016visualizing} or attention weights in some neural attention models~\citep{bahdanau2015neural}.
However, both of them are criticized about their ability of generating explanations~\citep{adebayo2018sanity,Jain2019AttentionIN}.

As mentioned in \autoref{sec:intro}, we differentiate the explainability of the local explanations from the local explanations themselves. 
As discussed in~\citep{gilpin2018explaining}, the explainability of an explanation is about whether it is easy for humans to interpret or understand. 
This distinction emphasizes the importance of human evaluation and helps us design the automatic evaluation method in this work.
For general discussion on explainability, we refer the readers to \citep{guidotti2018survey,gilpin2018explaining,murdoch2019interpretable}.

To improve the explainability of neural sentiment classifiers, we propose two data augmentation methods to add more training examples.
Prior work on data augmentation for text classification~\citep{kobayashi2018contextual,wei2019eda} mainly focuses more on improving prediction performance.
Adversarial examples, as a particular category of augmented data, are mainly used to enhance model robustness~\citep{li2017robust,sun2018training}.
With the two proposed methods, this work demonstrates a novel way of constructing and using augmented examples.

\section{Conclusion}
\label{sec:conclusion}
In this paper, we showed that the explainability of neural sentiment classifiers can be improved by training with augmented data.
We proposed two data augmentation methods by employing a predefined word list and adversarial examples respectively.
In this work, we focused on the explainability of local explanations, which were generated by LIME and the cosine similarity method. 
Then, the improvement of model explainability was assessed with both automatic evaluation and human evaluation.
Experiments showed that the proposed data augmentation methods could successfully improve the model explainability.

\small
\bibliography{ref}

\begin{thebibliography}{}

\bibitem[Adebayo et~al., 2018]{adebayo2018sanity}
Adebayo, J., Gilmer, J., Muelly, M., Goodfellow, I., Hardt, M., and Kim, B.
  (2018).
\newblock Sanity checks for saliency maps.
\newblock In {\em Advances in Neural Information Processing Systems}, pages
  9525--9536.

\bibitem[Alzantot et~al., 2018]{alzantot2018generating}
Alzantot, M., Sharma, Y., Elgohary, A., Ho, B.-J., Srivastava, M., and Chang,
  K.-W. (2018).
\newblock Generating natural language adversarial examples.
\newblock {\em arXiv preprint arXiv:1804.07998}.

\bibitem[Baccianella et~al., 2010]{baccianella2010sentiwordnet}
Baccianella, S., Esuli, A., and Sebastiani, F. (2010).
\newblock Sentiwordnet 3.0: an enhanced lexical resource for sentiment analysis
  and opinion mining.

\bibitem[Bahdanau et~al., 2015]{bahdanau2015neural}
Bahdanau, D., Cho, K., and Bengio, Y. (2015).
\newblock Neural machine translation by jointly learning to align and
  translate.
\newblock In {\em ICLR}.

\bibitem[Chelba et~al., 2013]{chelba2013one}
Chelba, C., Mikolov, T., Schuster, M., Ge, Q., Brants, T., Koehn, P., and
  Robinson, T. (2013).
\newblock One billion word benchmark for measuring progress in statistical
  language modeling.
\newblock {\em arXiv preprint arXiv:1312.3005}.

\bibitem[Chen et~al., 2016]{chen2016neural}
Chen, H., Sun, M., Tu, C., Lin, Y., and Liu, Z. (2016).
\newblock Neural sentiment classification with user and product attention.
\newblock In {\em Proceedings of the 2016 conference on empirical methods in
  natural language processing}, pages 1650--1659.

\bibitem[Chen et~al., 2018]{chen2018learning}
Chen, J., Song, L., Wainwright, M.~J., and Jordan, M.~I. (2018).
\newblock Learning to explain: An information-theoretic perspective on model
  interpretation.
\newblock {\em arXiv preprint arXiv:1802.07814}.

\bibitem[Dabkowski and Gal, 2017]{dabkowski2017real}
Dabkowski, P. and Gal, Y. (2017).
\newblock Real time image saliency for black box classifiers.
\newblock In {\em Advances in Neural Information Processing Systems}, pages
  6967--6976.

\bibitem[Daniel et~al., 2017]{daniel2017company}
Daniel, M., Neves, R.~F., and Horta, N. (2017).
\newblock Company event popularity for financial markets using twitter and
  sentiment analysis.
\newblock {\em Expert Systems with Applications}, 71:111--124.

\bibitem[Feldman, 2013]{feldman2013techniques}
Feldman, R. (2013).
\newblock Techniques and applications for sentiment analysis.
\newblock {\em Communications of the ACM}, 56(4):82--89.

\bibitem[Fong and Vedaldi, 2017]{fong2017interpretable}
Fong, R.~C. and Vedaldi, A. (2017).
\newblock Interpretable explanations of black boxes by meaningful perturbation.
\newblock In {\em Proceedings of the IEEE International Conference on Computer
  Vision}, pages 3429--3437.

\bibitem[Gilpin et~al., 2018]{gilpin2018explaining}
Gilpin, L.~H., Bau, D., Yuan, B.~Z., Bajwa, A., Specter, M., and Kagal, L.
  (2018).
\newblock Explaining explanations: An overview of interpretability of machine
  learning.
\newblock In {\em 2018 IEEE 5th International Conference on Data Science and
  Advanced Analytics (DSAA)}, pages 80--89. IEEE.

\bibitem[Goodfellow et~al., 2014]{goodfellow2014explaining}
Goodfellow, I.~J., Shlens, J., and Szegedy, C. (2014).
\newblock Explaining and harnessing adversarial examples.
\newblock {\em arXiv preprint arXiv:1412.6572}.

\bibitem[Guidotti et~al., 2018]{guidotti2018survey}
Guidotti, R., Monreale, A., Ruggieri, S., Turini, F., Giannotti, F., and
  Pedreschi, D. (2018).
\newblock A survey of methods for explaining black box models.
\newblock {\em ACM computing surveys (CSUR)}, 51(5):93.

\bibitem[Hochreiter and Schmidhuber, 1997]{hochreiter1997long}
Hochreiter, S. and Schmidhuber, J. (1997).
\newblock Long short-term memory.
\newblock {\em Neural computation}, 9(8):1735--1780.

\bibitem[Jain and Wallace, 2019]{Jain2019AttentionIN}
Jain, S. and Wallace, B.~C. (2019).
\newblock Attention is not explanation.
\newblock {\em CoRR}, abs/1902.10186.

\bibitem[Johnson and Zhang, 2016]{johnson2016supervised}
Johnson, R. and Zhang, T. (2016).
\newblock Supervised and semi-supervised text categorization using lstm for
  region embeddings.
\newblock {\em arXiv preprint arXiv:1602.02373}.

\bibitem[Johnson and Zhang, 2017]{johnson2017deep}
Johnson, R. and Zhang, T. (2017).
\newblock Deep pyramid convolutional neural networks for text categorization.
\newblock In {\em Proceedings of the 55th Annual Meeting of the Association for
  Computational Linguistics (Volume 1: Long Papers)}, pages 562--570.

\bibitem[Kim, 2014]{kim2014convolutional}
Kim, Y. (2014).
\newblock Convolutional neural networks for sentence classification.
\newblock {\em arXiv preprint arXiv:1408.5882}.

\bibitem[Kindermans et~al., 2017]{kindermans2017learning}
Kindermans, P.-J., Sch{\"u}tt, K.~T., Alber, M., M{\"u}ller, K.-R., Erhan, D.,
  Kim, B., and D{\"a}hne, S. (2017).
\newblock Learning how to explain neural networks: Patternnet and
  patternattribution.
\newblock {\em arXiv preprint arXiv:1705.05598}.

\bibitem[Kingma and Ba, 2014]{kingma2014adam}
Kingma, D.~P. and Ba, J. (2014).
\newblock Adam: A method for stochastic optimization.
\newblock {\em arXiv preprint arXiv:1412.6980}.

\bibitem[Kobayashi, 2018]{kobayashi2018contextual}
Kobayashi, S. (2018).
\newblock Contextual augmentation: Data augmentation by words with paradigmatic
  relations.
\newblock {\em arXiv preprint arXiv:1805.06201}.

\bibitem[Kurakin et~al., 2016]{kurakin2016adversarial}
Kurakin, A., Goodfellow, I., and Bengio, S. (2016).
\newblock Adversarial machine learning at scale.
\newblock {\em arXiv preprint arXiv:1611.01236}.

\bibitem[Lei et~al., 2016]{lei2016rationalizing}
Lei, T., Barzilay, R., and Jaakkola, T. (2016).
\newblock Rationalizing neural predictions.
\newblock {\em arXiv preprint arXiv:1606.04155}.

\bibitem[Li et~al., 2016]{li2016visualizing}
Li, J., Chen, X., Hovy, E., and Jurafsky, D. (2016).
\newblock Visualizing and understanding neural models in nlp.
\newblock In {\em Proceedings of NAACL-HLT}, pages 681--691.

\bibitem[Li et~al., 2018]{li2018transformation}
Li, X., Bing, L., Lam, W., and Shi, B. (2018).
\newblock Transformation networks for target-oriented sentiment classification.
\newblock {\em arXiv preprint arXiv:1805.01086}.

\bibitem[Li et~al., 2017]{li2017robust}
Li, Y., Cohn, T., and Baldwin, T. (2017).
\newblock Robust training under linguistic adversity.
\newblock In {\em Proceedings of the 15th Conference of the European Chapter of
  the Association for Computational Linguistics: Volume 2, Short Papers}, pages
  21--27.

\bibitem[Liu, 2012]{liu2012sentiment}
Liu, B. (2012).
\newblock Sentiment analysis and opinion mining.
\newblock {\em Synthesis lectures on human language technologies}, 5(1):1--167.

\bibitem[Lundberg and Lee, 2017]{lundberg2017unified}
Lundberg, S.~M. and Lee, S.-I. (2017).
\newblock A unified approach to interpreting model predictions.
\newblock In {\em Advances in Neural Information Processing Systems}, pages
  4765--4774.

\bibitem[Luo et~al., 2018]{luo2018beyond}
Luo, L., Ao, X., Pan, F., Wang, J., Zhao, T., Yu, N., and He, Q. (2018).
\newblock Beyond polarity: Interpretable financial sentiment analysis with
  hierarchical query-driven attention.
\newblock In {\em IJCAI}.

\bibitem[Maas et~al., 2011]{maas2011learning}
Maas, A.~L., Daly, R.~E., Pham, P.~T., Huang, D., Ng, A.~Y., and Potts, C.
  (2011).
\newblock Learning word vectors for sentiment analysis.
\newblock In {\em Proceedings of the 49th annual meeting of the association for
  computational linguistics: Human language technologies-volume 1}, pages
  142--150. Association for Computational Linguistics.

\bibitem[Madry et~al., 2017]{madry2017towards}
Madry, A., Makelov, A., Schmidt, L., Tsipras, D., and Vladu, A. (2017).
\newblock Towards deep learning models resistant to adversarial attacks.
\newblock {\em arXiv preprint arXiv:1706.06083}.

\bibitem[Mikolov et~al., 2013]{mikolov2013distributed}
Mikolov, T., Sutskever, I., Chen, K., Corrado, G.~S., and Dean, J. (2013).
\newblock Distributed representations of words and phrases and their
  compositionality.
\newblock In {\em Advances in neural information processing systems}, pages
  3111--3119.

\bibitem[Murdoch et~al., 2019]{murdoch2019interpretable}
Murdoch, W.~J., Singh, C., Kumbier, K., Abbasi-Asl, R., and Yu, B. (2019).
\newblock Interpretable machine learning: definitions, methods, and
  applications.
\newblock {\em arXiv preprint arXiv:1901.04592}.

\bibitem[Nguyen, 2018]{nguyen2018comparing}
Nguyen, D. (2018).
\newblock Comparing automatic and human evaluation of local explanations for
  text classification.
\newblock In {\em Proceedings of the 2018 Conference of the North American
  Chapter of the Association for Computational Linguistics: Human Language
  Technologies, Volume 1 (Long Papers)}, pages 1069--1078.

\bibitem[Pang and Lee, 2005]{pang2005seeing}
Pang, B. and Lee, L. (2005).
\newblock Seeing stars: Exploiting class relationships for sentiment
  categorization with respect to rating scales.
\newblock In {\em Proceedings of the 43rd annual meeting on association for
  computational linguistics}, pages 115--124. Association for Computational
  Linguistics.

\bibitem[Pang et~al., 2002]{pang2002thumbs}
Pang, B., Lee, L., and Vaithyanathan, S. (2002).
\newblock Thumbs up?: sentiment classification using machine learning
  techniques.
\newblock In {\em Proceedings of the ACL-02 conference on Empirical methods in
  natural language processing-Volume 10}, pages 79--86. Association for
  Computational Linguistics.

\bibitem[Ribeiro et~al., 2016]{ribeiro2016should}
Ribeiro, M.~T., Singh, S., and Guestrin, C. (2016).
\newblock Why should i trust you?: Explaining the predictions of any
  classifier.
\newblock In {\em Proceedings of the 22nd ACM SIGKDD international conference
  on knowledge discovery and data mining}, pages 1135--1144. ACM.

\bibitem[Ribeiro et~al., 2018]{ribeiro2018anchors}
Ribeiro, M.~T., Singh, S., and Guestrin, C. (2018).
\newblock Anchors: High-precision model-agnostic explanations.
\newblock In {\em Thirty-Second AAAI Conference on Artificial Intelligence}.

\bibitem[Socher et~al., 2013]{socher2013recursive}
Socher, R., Perelygin, A., Wu, J., Chuang, J., Manning, C.~D., Ng, A., and
  Potts, C. (2013).
\newblock Recursive deep models for semantic compositionality over a sentiment
  treebank.
\newblock In {\em Proceedings of the 2013 conference on empirical methods in
  natural language processing}, pages 1631--1642.

\bibitem[Sohangir et~al., 2018]{sohangir2018big}
Sohangir, S., Wang, D., Pomeranets, A., and Khoshgoftaar, T.~M. (2018).
\newblock Big data: Deep learning for financial sentiment analysis.
\newblock {\em Journal of Big Data}, 5(1):3.

\bibitem[Sun et~al., 2018]{sun2018training}
Sun, S., Yeh, C.-F., Ostendorf, M., Hwang, M.-Y., and Xie, L. (2018).
\newblock Training augmentation with adversarial examples for robust speech
  recognition.
\newblock {\em arXiv preprint arXiv:1806.02782}.

\bibitem[Wei and Zou, 2019]{wei2019eda}
Wei, J.~W. and Zou, K. (2019).
\newblock Eda: Easy data augmentation techniques for boosting performance on
  text classification tasks.
\newblock {\em arXiv preprint arXiv:1901.11196}.

\bibitem[Yang et~al., 2016]{yang2016hierarchical}
Yang, Z., Yang, D., Dyer, C., He, X., Smola, A., and Hovy, E. (2016).
\newblock Hierarchical attention networks for document classification.
\newblock In {\em Proceedings of the 2016 Conference of the North American
  Chapter of the Association for Computational Linguistics: Human Language
  Technologies}, pages 1480--1489.

\bibitem[Yu and Kumbier, 2019]{yu2019three}
Yu, B. and Kumbier, K. (2019).
\newblock Three principles of data science: predictability, computability, and
  stability (pcs).
\newblock {\em arXiv preprint arXiv:1901.08152}.

\bibitem[Zhang, 2015]{zhang2015incorporating}
Zhang, Y. (2015).
\newblock Incorporating phrase-level sentiment analysis on textual reviews for
  personalized recommendation.
\newblock In {\em Proceedings of the eighth ACM international conference on web
  search and data mining}, pages 435--440. ACM.

\bibitem[Zhang et~al., 2014]{zhang2014explicit}
Zhang, Y., Lai, G., Zhang, M., Zhang, Y., Liu, Y., and Ma, S. (2014).
\newblock Explicit factor models for explainable recommendation based on
  phrase-level sentiment analysis.
\newblock In {\em Proceedings of the 37th international ACM SIGIR conference on
  Research \& development in information retrieval}, pages 83--92. ACM.

\bibitem[Zhou et~al., 2016]{zhou2016text}
Zhou, P., Qi, Z., Zheng, S., Xu, J., Bao, H., and Xu, B. (2016).
\newblock Text classification improved by integrating bidirectional lstm with
  two-dimensional max pooling.
\newblock {\em arXiv preprint arXiv:1611.06639}.

\end{thebibliography}
\bibliographystyle{apalike}

\clearpage
\appendix
\section{Implementation Details}
\label{sec:other}

\paragraph{Datasets} We use three sentiment benchmark datasets for evaluation. Summary statistics of the datasets are in \autoref{tab:datasets} .

\begin{table}
  \centering
  \begin{tabular}{p{0.15\textwidth}lll}
    	\toprule
         Dataset & \textit{C} & \textit{L} & \textit{N} \\
         \midrule
         SST & 2 & 19 & 9613 \\
         MR & 2 & 20 & 10662 \\
         IMDB & 2 & 268 & 50000 \\
         \bottomrule
  \end{tabular}
  \caption{Summary statistics for the datasets, where \textit{C} is the number of classes, \textit{L} is average sentence length, and \textit{N} is dataset size.}
  \label{tab:datasets}
\end{table}

\begin{itemize}
	\item \textbf{SST.} This dataset was proposed in \citep{socher2013recursive} for sentence-level sentiment classification. We used the SST-2, which is the 2-class version of this dataset. There are 6,920 examples in the training set, 872 examples in the development set, and 1,821 examples in the test set. For data augmentation with \daek, additional 1,624 and 229 augmented examples were added to the training and development sets respectively. With \daadv, 4,885 and 539 augmented examples were added to the training and development sets respectively.
	
	\item \textbf{MR.} This dataset was proposed by \citet{pang2005seeing}. These reviews in this dataset were divided into 9,596 training and 1066 test examples. In our experiments, 90\% of the training examples are used for training, and the rest is used as development set. With \daek, we added additional 4,318 and 480 augmented examples to the training and development sets respectively.
	
	\item \textbf{IMDB.} This dataset was proposed by \citet{maas2011learning}. These reviews in this dataset were divided into 25,000 training and 25,000 test examples. We split 90\% of the training examples for training, and the rest as the development set. With \daek, additional 11,250  and 1,250 augmented examples were added to the training and development sets respectively.
\end{itemize}

\paragraph{Neural Sentiment Classifiers} In this work, we use  a convolutional neural network in \citep[\oldmodel]{kim2014convolutional} and a recurrent neural network with LSTM \citep[\oldrnn]{hochreiter1997long} as our baseline models.

The \oldmodel consists of an input layer that takes word embeddings as inputs, a convolutional layer followed by a max-pooling layer for composing word embeddings into text representations, and a softmax layer for classification.
For a given text $\vec{x}$, $\vec{f}(\cdot)$ denotes the representation function in \oldmodel, which maps $\vec{x}$ into a $d$-dimensional numeric vector $\vec{f}(\vec{x})$ as text representation.
The decision function is defined as
\begin{equation}
\label{eq:cnn-decision}
\vec{h}(\vec{x},y) = \trans{\vec{u}_y}\vec{f}(\vec{x}),
\end{equation}
where $y\in\mathcal{Y}$ is the class label, and $\mat{u}_y\in\mathbb{R}^{d}$ is the corresponding classification weight vector.
For prediction, we use $\hat{y}=\argmax_{y}\vec{h}(\vec{x},y)$.

The \oldrnn consists of uni-directional one-layer LSTM.
For a given text, the last hidden state of this RNN is used as the text representation $\vec{f}(\vec{x})$.
The same decision function defined in \autoref{eq:cnn-decision} is employed for sentiment prediction. 

Even though the main focus of this work is on model explainability, the prerequisite is to match the classification performance in prior work with the similar model architectures.
Here are some implementation details that we adopted from prior work \citep{kim2014convolutional}:
For both the \oldmodel, \newmodel and \newmodeladv, we used a single convolutional layer with filters of the window sizes ranging from 3 to 5. For both \oldrnn and \newrnn, we used a single layer LSTM. For all of the models, the input parameters were initialized with the 300-dimensional pretrained word embeddings \citep[word2vec]{mikolov2013distributed} and all other parameters were randomly initialized with the default method in PyTorch.
Hyperparameters, including kernel size (for CNN only), hidden size (for RNN only), learning rate, minibatch size, etc., were tuned separately on the development set for different datasets.
We used Adam \citep{kingma2014adam} to update the parameters. 

\paragraph{LIME Setup} For LIME, the number of subsamples was set to 600, $\sigma^{2}$ was $10$, and each local linear model was trained over 50 epochs. For adversarial training, the attacker is limited to 20 iterations, and some other hyper-parameters are fixed: nearest neighbors was 8,  top kept words was 4, and the maximum percentage of allowed change to the text was 20\%.

\end{document}